\pdfoutput=1

\documentclass[11pt]{article}
\usepackage{amsfonts}
\usepackage{EMNLP2022}
\usepackage{graphicx}
\usepackage{times}
\usepackage{latexsym}
\usepackage{placeins}

\usepackage[T1]{fontenc}

\usepackage[utf8]{inputenc}

\usepackage{adjustbox}

\usepackage{microtype}

\usepackage{inconsolata}
\usepackage{comment}

\title{Exploring Hybrid and Ensemble Models for Multiclass Prediction of Mental Health Status on Social Media}

\author{Sourabh Zanwar \\
  RWTH Aachen University \\
  \texttt{sourabh.zanwar@rwth-aachen.de} \\\And
  Daniel Wiechmann \\
  University of Amsterdam \\
  \texttt{d.wiechmann@uva.nl} \\ \AND
  Yu Qiao \\
  RWTH Aachen University \\
  \texttt{yu.qiao@rwth-aachen.de} \\\And
  Elma Kerz \\
  RWTH Aachen University \\
  \texttt{elma.kerz@ifaar.rwth-aachen.de} \\
  }

\begin{document}
\maketitle
\begin{abstract}
In recent years, there has been a surge of interest in research on automatic mental health detection (MHD) from social media data leveraging advances in natural language processing and machine learning techniques. While significant progress has been achieved in this interdisciplinary research area, the vast majority of work has treated MHD as a binary classification task.  The multiclass classification setup is, however, essential if we are to uncover the subtle differences among the statistical patterns of language use associated with particular mental health conditions. Here, we report on experiments aimed at predicting six conditions (anxiety, attention deficit hyperactivity disorder, bipolar disorder, post-traumatic stress disorder, depression, and psychological stress) from Reddit social media posts. We explore and compare the performance of hybrid and ensemble models leveraging transformer-based architectures (BERT and RoBERTa) and BiLSTM neural networks trained on within-text distributions of a diverse set of linguistic features. This set encompasses measures of syntactic complexity, lexical sophistication and diversity, readability, and register-specific ngram frequencies, as well as sentiment and emotion lexicons. In addition, we conduct feature ablation experiments to investigate which types of features are most indicative of particular mental health conditions.

\end{abstract}

\section{Introduction}

Mental health is a major challenge in healthcare and in our modern societies at large, as evidenced by the topic's inclusion in the United Nations' 17 Sustainable Development Goals. The World Health Organization estimates that 970 million people worldwide suffer from mental health issues, the most common being anxiety and depressive disorders\footnote{\url{https://www.who.int/news-room/fact-sheets/detail/mental-disorders}}.  The problem is compounded by the fact that the rate of undiagnosed mental disorders has been estimated to be as high as 45\%
\cite{la2012undiagnosed}.  The societal impact of mental health disorders requires prevention and intervention strategies focused primarily on screening and early diagnosis. In keeping with the WHO Mental Health Action Plan \cite{saxena2013world}, natural language processing and machine learning can make an important contribution to gathering more comprehensive information and knowledge about mental illness. In particular, an increasing use of social media platforms by individuals is generating large amounts of high-quality behavioral and textual data that can support the development of computational solutions for the study of mental disorders. An emerging, interdisciplinary field of research at the intersections of computational linguistics, health informatics and artificial intelligence now leverages natural language processing techniques to analyze such data to develop models for early detection of various mental health conditions.

Systematic reviews of this research show that the vast majority of the existing work has focused primarily on automatic identification of specific disorders, with depression and anxiety being the most commonly studied target conditions \cite{calvo2017natural,chancellor2020methods,zhang2022natural}. As a result, existing  work has focused on developing binary classifiers that aim to distinguish between individuals with a particular mental illness and control users. 

The current work addresses the more complex problem of distinguishing between multiple mental states, which is essential if we are to uncover the subtle differences among the statistical patterns of language use associated with particular disorders. Specifically, in this paper we make the following contributions to the existing literature on health text mining based on social media data: (1) We frame the MHC detection tasks as a multiclass prediction task aimed to determine to what extent six  mental health conditions (anxiety, attention deficit hyperactivity disorder, bipolar disorder, post-traumatic stress disorder, depression, and psychological stress) can be predicted on the basis of social media posts from Reddit. (2) We explore and compare the performance of hybrid and ensemble models leveraging transformer-based architectures (BERT and RoBERTa) and BiLSTM neural networks trained on within-text distributions of a diverse set of linguistic features. (3) We conduct feature ablation experiments to investigate which types of features are most indicative of particular mental health conditions.

This paper is organized into five sections. Section \ref{related} provides a concise overview of the current state of research on mental health detection from Reddit social media posts. Section \ref{exp} presents the experimental setup including descriptions of the data, the type of linguistic features used and their computation, and the modeling approach. The main results are presented and discussed in Section \ref{results}. In Section \ref{conclusion} general conclusions are drawn and an outlook is given.

\section{Related work} \label{related}

A growing body of research has demonstrated that NLP techniques in combination with text data from social media provide a valuable approach to understanding and modeling people's mental health and have the potential to enable more individualized and scalable methods for timely mental health care (see \citet{calvo2017natural, chancellor2020methods, zhang2022natural}, for systematic reviews). A surge in the number of research initiatives by way of workshops and shared tasks, such as  Computational Linguistics and Clinical Psychology (CLPsych) Workshop, Social Media Mining for Health Applications (SMMH) and International Workshop on Health Text Mining and Information Analysis (LOUHI), are advancing this research area: It fosters an interdisciplinary approach to automatic methods for the collection, extraction, representation, and analysis of social media data for health informatics and text mining that tightly integrates insights from clinical and cognitive psychology with natural language processing and machine learning. It actively contributes to making publicly available large labeled and high quality datasets, the availability of which has a significant impact on modeling and understanding mental health.

While earlier research on social media mining for health applications has been conducted primarily with Twitter texts \cite{braithwaite2016validating,coppersmith2014quantifying}, a more recent stream of research has turned towards leveraging Reddit as a richer source for constructing mental health benchmark datasets \cite{cohan2018smhd,turcan-mckeown-2019-dreaddit}. Reddit is an interactive, discussion-oriented platform without any length constraints like Twitter, where posts are limited to 280 characters.  Its users, the Redditors, are anonymous and the site is clearly organized into more than two million different topics, subreddits. Another crucial fact that makes Reddit more suitable for health text mining is that, unlike Twitter (with its limited text length), extended text production provides a richer linguistic signal that allows analysis at all levels of organization (morpho-syntactic complexity, lexical and phrasal variety, and sophistication and readability). \citet{yates-etal-2017-depression}, for instance, proposed an approach for automatically labeling the mental health status of Reddit users. Reflecting the topic organization of Reddits with its subreddits, the authors created high precision patterns to identify users who claimed to have been diagnosed with a mental health condition (diagnosed users) and used exclusion criteria to match them with control users. To prevent easy identification of diagnosed users, the resulting dataset excluded all obvious expressions used to construct it. This approach was also adapted to other mental health conditions \cite{cohan2018smhd}. 

Previous research on health text mining from social media posts has primarily focused on the automatic identification of specific mental disorders and has treated it as a binary classification task aimed at distinguishing between users with a target mental condition and control ones (see the systematic reviews mentioned above). To the best of our knowledge, the only two exceptions are \citet{gkotsis2017characterisation} and \citet{murarka2021classification}. \citet{gkotsis2017characterisation} proposed an approach to classify mental health-related posts according to theme-based subreddit groupings using deep learning techniques. The authors constructed a dataset of 458,240 posts from mental health related subreddits paired with a control set approximately matched in size (476,388 posts).  The mental health-related posts were grouped into 11 MHC themes (addiction, autism, anxiety, bipolar, BPD, depression, schizophrenia, selfharm, SuicideWatch, cripplingalcoholism, opiates) based on a combination of manual assessment steps and automated topic detection. Their best performing model, a convolutional neural network classifier trained on word embeddings, was able to identify the correct theme with a weighted average accuracy of 71.37\%. The approach taken in this work was primarily aimed at identifying posts that are relevant to a mental health subreddit, as well as the actual mental health topic to which they relate. Another more recent exception similar to our work is \citet{murarka2021classification}. The authors used RoBERTa (Robustly Optimized BERT Pretraining Approach, \citet{liu2019roberta}) to build multiclass models to identify five mental health conditions from Reddit posts (ADHD, anxiety, bipolar disorder, depression, and PTSD).  The model was trained on a dataset consisting of Reddit subreddits with 17,159 posts. The RoBERTa-based model achieved a macro-averaged F1 value of 89\%, with F1 values for individual conditions ranging from 84\% for depression to 91\% for ADHD. Although these results appear impressive, they should be interpreted with caution: To obtain data for each of the mental health conditions, the authors extracted posts from five subreddits (r/adhd, r/anxiety, r/bipolar r/disorder, r/depression, r/ptsd) and assigned them a class label corresponding to the name of the condition with which they were associated. Posts for the control group were selected from subreddits with a wide range of general topics (music, travel, India, politics, English, datasets, mathematics and science). The way the datasets in \citet{gkotsis2017characterisation} and \citet{murarka2021classification} are constructed rendered the classification tasks relatively easy, as it allows the classifier to use explicit mentions of mental health terms associated with a particular mental health condition. However, there is growing recognition that careful dataset construction is critical to developing robust and generalizable models for detecting mental health status on social media. This requires the removal of expressions indicating mental health status for both diagnosed and control users  (see \cite{yates-etal-2017-depression} or SMHD \cite{cohan2018smhd}; see also \citet{chancellor2020methods} and \citet{harrigian2021state} for discussions on obtaining ground truth labels for the positive classes and data preprocessing/selection).

The existing research on the detection of mental health conditions in social media mainly follows one of two approaches: One focuses on linguistic features, mainly in the form of unigrams with TF-IDF (term frequency-inverse document frequency)  weighting, or on specialized dictionaries, especially the categories from the Linguistic Inquiry and Word Count (LIWC) dictionaries \cite{de2013social,nguyen2014affective,sekulic-strube-2019-adapting,zomick2019linguistic}.  The second centers on leveraging contextualized embedding techniques and pre-trained language models such as BERT \cite{devlin2018bertN}, ELMo \cite{peters-etal-2018-deep}, and RoBERTa \cite{liu2019RoBERTa}, minimizing the need for tasks such as feature engineering or feature selection (\citet{gkotsis2017characterisation,murarka2021classification}, see also \citet{su2020deep} for a review). However, less work has been undertaken to date to explore hybrid and ensemble models for mental illness recognition that integrate engineered features with transformer-based language models. Such hybrid models have recently been successfully applied in the neighboring research area of personality recognition \cite{mehta2020bottom,kerz2022pushing}.

\section{Experimental setup} \label{exp}

\subsection{Dataset} \label{data}

The dataset used in this work was constructed from two recent corpora used for the detection of MHC: (1) the Self-Reported Mental Health Diagnoses (SMHD) dataset  \cite{cohan2018smhd} and (2) the Dreaddit dataset \cite{turcan-mckeown-2019-dreaddit}. Both SMHD and Dreaddit were compiled from Reddit, a social media platform consisting of individual topic communities called subreddits, including those relevant to MHC detection. The length of Reddit posts makes them a particularly valuable resource, as it allows modeling of the distribution of linguistic features in the text. 

SMHD is a large dataset of social media posts from users with nine mental health conditions (MHC) corresponding to branches in the DSM-5 \cite{edition2013diagnostic}, an authoritative taxonomy for psychiatric diagnoses. User-level MHC labels were obtained through carefully designed distantly supervised labeling processes based on diagnosis pattern matching. The pattern matching leveraged a seed list of diagnosis keywords collected from the corresponding DSM-5 headings and extended by synonym mappings. To prevent that target labels can be easily inferred from the presence of MHC indicating words/phrases in the posts, all posts made to mental health-related subreddits or containing keywords related to a mental health condition were removed from the diagnosed users' data. Dreaddit is a dataset of lengthy social media posts from subreddits in five domains that include stressful and non-stressful text. For a subset of 3.5k users employed in this paper, binary labels (+/- stressful) were obtained from aggregated ratings of five crowdsourced human annotators. 

Based on these two corpora, we constructed a dataset with the goal of obtaining sub-corpora of equal size for the six MHCs targeted in this paper. To this end, we downsampled SMHD to match the size of Dreaddit and to be balanced in terms of class distributions. The sampling procedure from the SMHD dataset was such that each post was produced by a distinct user. In doing so, we addressed a concerning trend described in recent review articles that points to the presence of a relatively small number of unique individuals, which may hinder the generalization of models to platforms that are already demographically skewed \cite{chancellor2020methods,harrigian2021state}. These constraints were met for five of the nine MHC in the SMHD dataset (attention deficit hyperactivity disorder (ADHD), anxiety, bipolar, depression, post-traumatic stress disorder (PTSD)). The data for the control groups contained the full Dreaddit control subset, which comtains just under 1700 posts, plus an additional 1805 control posts from the SMHD dataset that were matched in terms of post length. The control subset was intentionally designed as a majority class to reduce false positive (overdiagnosis) rates (see \citet{merten2017overdiagnosis} for discussion). Statistics for these datasets are presented in Table \ref{tab:dataStats}. 

\begin{table}
\caption{Datasets statistics (number of posts, means and standard deviations of  post length (in words) across mental health conditions and control groups.}
\label{tab:dataStats}
\setlength{\tabcolsep}{2.5pt}
\adjustbox{max width=0.9\textwidth}{
\begin{tabular}{lcccc}
\hline
MHC        & Dataset  & N posts & M length & SD   \\
\hline
Stress     & Dreaddit & 1857    & 91       & 35   \\
\hline
ADHD       & SMHD     & 1849    & 91.4     & 57   \\
Anxiety    & SMHD     & 1846    & 91.7     & 56.3 \\
Bipolar    & SMHD     & 1848    & 93       & 57.7 \\
Depression & SMHD     & 1846    & 92.4     & 58.7 \\
PTSD       & SMHD     & 1600    & 95.7     & 59.9 \\
\hline
Control    & Dreaddit & 1696    & 83.6     & 29.7 \\
    & SMHD     & 1805    & 78.8     & 48.6 \\
\hline
\end{tabular}
}
\vspace{-5mm}
\end{table}

\subsection{Measurement of within-text distributions of engineered features} \label{contours}

\begin{figure*}[h!]
    \centering
    \includegraphics[width=0.95\textwidth]{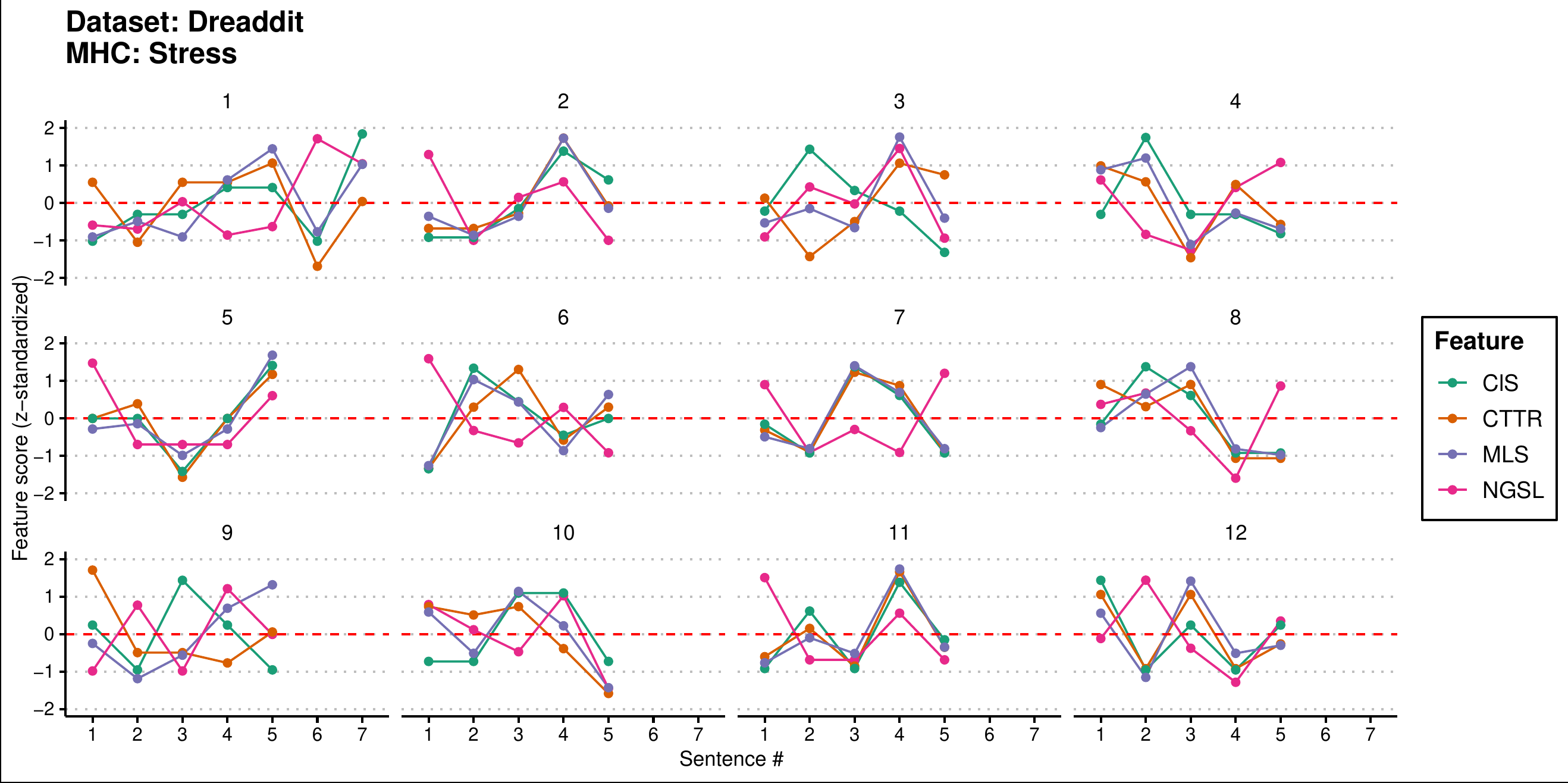}    \includegraphics[width=0.95\textwidth]{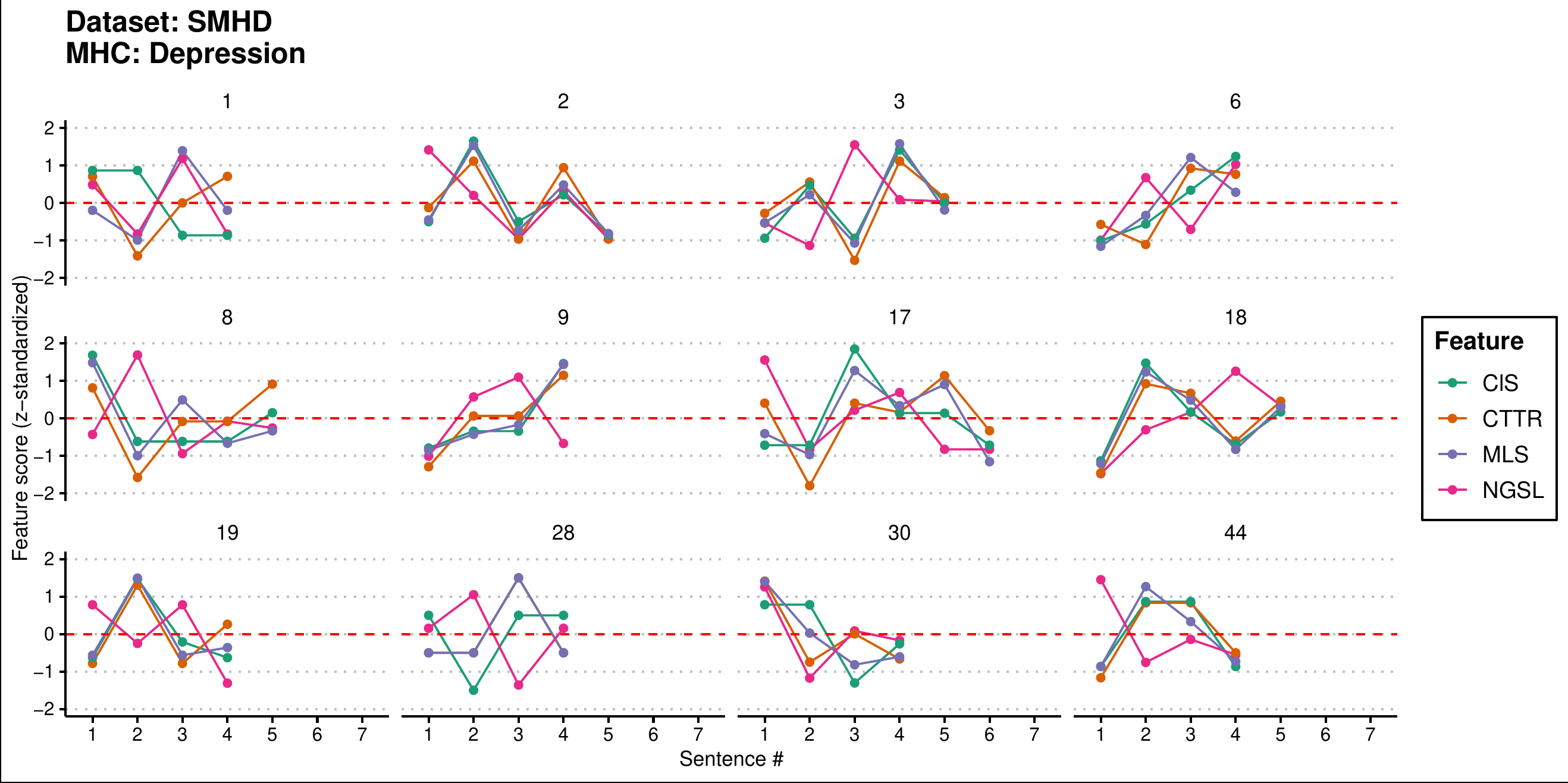}
    \caption{Within-text distributions of ClS (Clauses per Sentence), CTTR (corrected Type/Token Ratio), MLS (Mean Length of Sentence in Words), NGSL (Number of Sophisticated Words). Panels on top show the within-text distributions for 12 randomly selected Reddit posts categorized as exhibiting stress from the Dreaddit dataset. Bottom panels show the within-text distributions for 12 randomly selected posts from the SMHD daatset from users diagnosed with depression.}
    \label{fig:contours}
    \vspace{-5mm}
\end{figure*}

A diverse set of features used in this work fall into the following eight broad categories: (1) features of morpho-syntactic complexity (N=19), (2) features of lexical richness (N=52), (3) register-based n-gram frequency features (N=25), (4) readability features (N=14), and lexicon features designed to detect sentiment, emotion and/or affect (N=325). These features were subdivided into four categories: (5) Emotion/Sentiment, (6) LIWC, (7) Affect, and (8) General Inquirer. An overview of these features can be found in Table \ref{tab:features} in the appendix. All measurements of these features were calculated using an automated text analysis (ATA) system that employs a sliding window technique to compute sentence-level measurements (for recent applications of the ATA system in the context of text classification, see \citet{qiao21_interspeech} and \citet{kerz2022pushing}). These measurements capture the within-text distributions of scores for a given feature. Tokenization, sentence splitting, part-of-speech tagging, lemmatization and syntactic PCFG parsing were performed using Stanford CoreNLP \citep{manning2014stanford}. 

Figure \ref{fig:contours} provides \textbf{some examples} of within-text distributions for four selected features for twelve randomly selected Reddit posts from two datasets used in our work. Each of panels in Figure \ref{fig:contours} shows the distributions of four of the 436 textual features for one 24 randomly selected texts. The panels on top show the within-text distributions for 12 randomly selected Reddit posts categorized as exhibiting stress from the Dreaddit dataset. The panels panels on the bottom show the within-text distributions for 12 randomly selected posts from the SMHD daatset from users diagnosed with depression. We note that the distribution of feature values is generally not uniform, but shows large fluctuations over the course of the text. Furthermore, high values in one feature are often counterbalanced by low values in another feature. The classification models described in Section 3.3 are designed to detect local peaks of particular features and exploit the fluctuations for the detection of specific MHCs.

\subsection{Modeling approach}
\begin{figure}
    \centering
    \includegraphics[width = 0.3\textwidth]{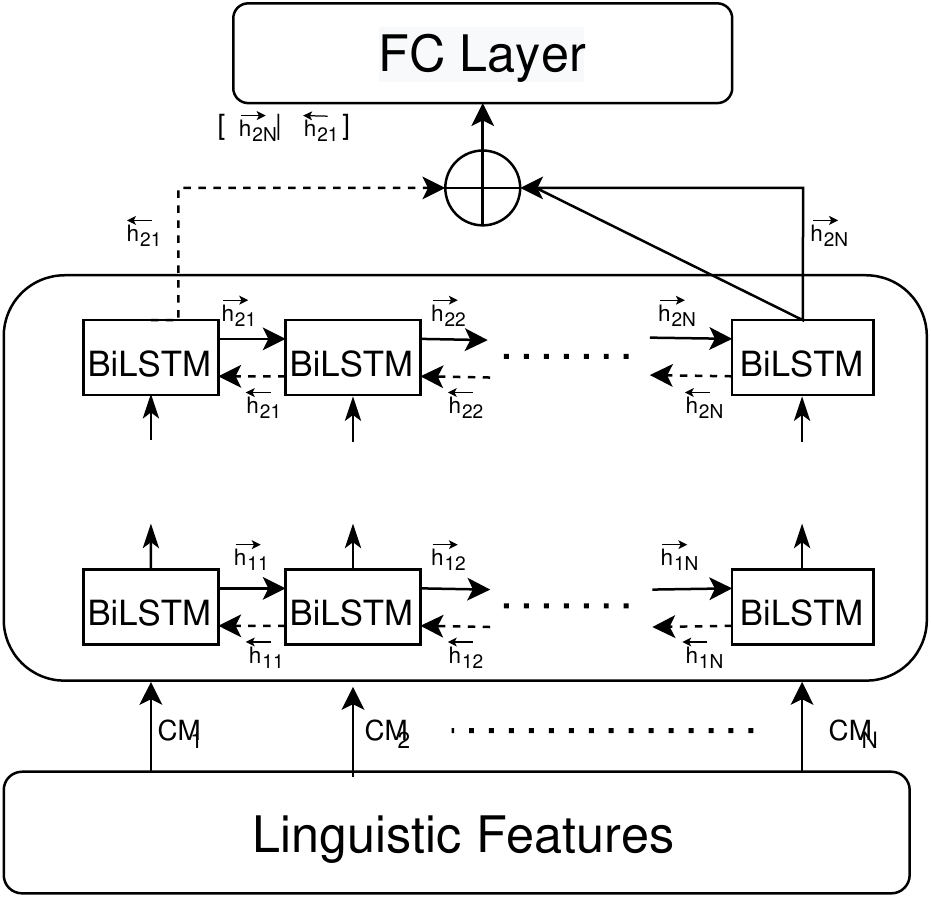}
    \caption{Structure diagram of BiLSTM mental health classification model trained on linguistic features}
    \label{fig:strucPsy}
\end{figure}

\begin{figure}
    \centering
    \includegraphics[width = 0.45\textwidth]{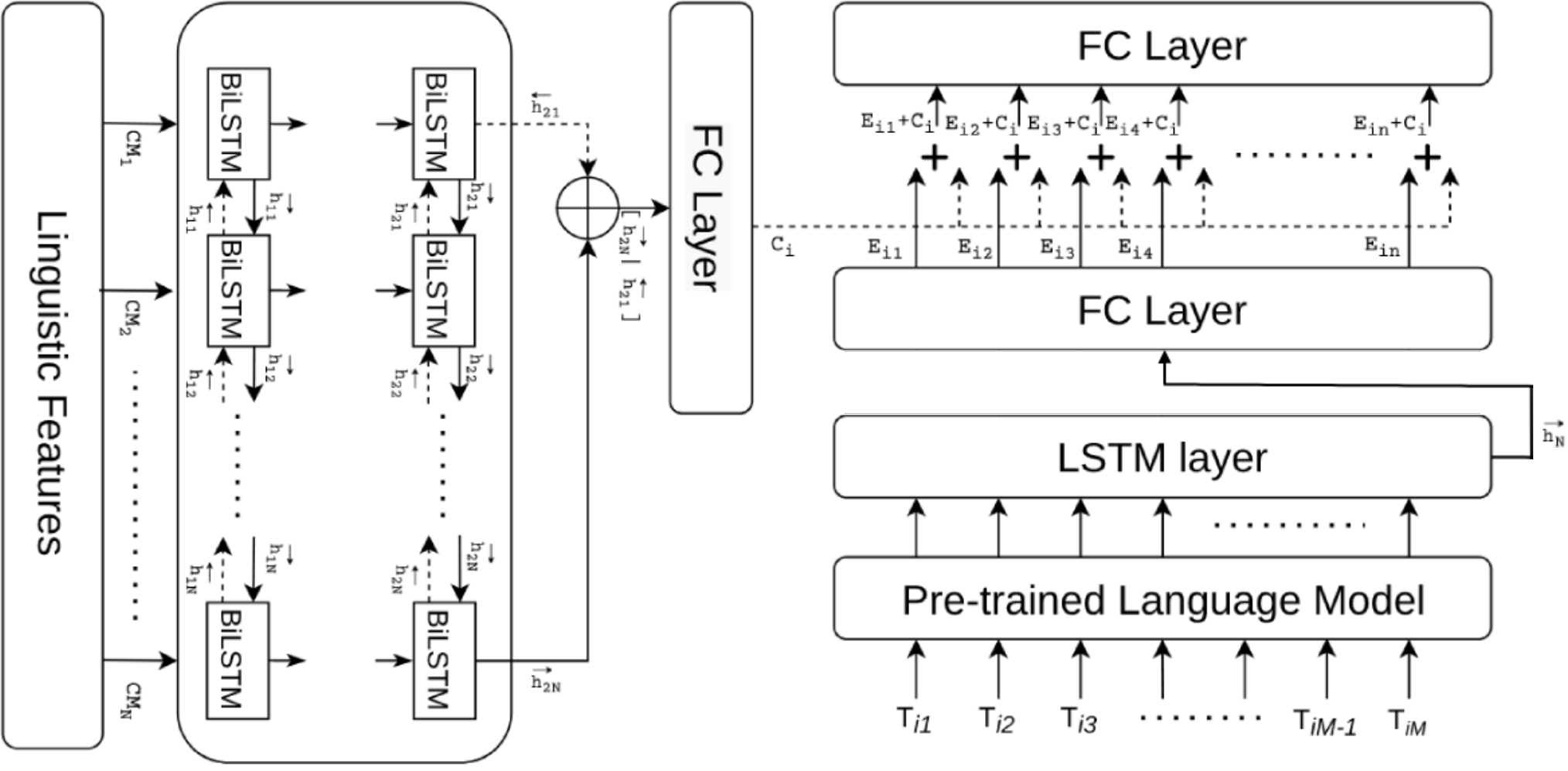}
    \caption{Structure diagram of the hybrid mental health classification models}
    \label{fig:strucHyb}
\end{figure}

We built five multiclass classification models to predict six mental health conditions (depression, anxiety, bipolar, ADHD, stress and PTSD): Two of these models leverage transformer-based architectures: BERT \cite{devlin2018bertN} and RoBERTa \cite{zhuang2019roberta}. These serve as the baseline models and components of our hybrid model. We used the pretrained `bert-base-uncased' and `roberta-base' models from the Huggingface Transformers library \cite{wolf2020Transformers}, each with an intermediate bidirectional long short-term memory (BiLSTM) layer with 256 hidden units \cite{9078946}. The third model is a BiLSTM classifier (Psyling-BiLSTM) trained solely on the eight feature groups described in Section \ref{contours}. Specifically, we constructed a 4-layer BiLSTM with a hidden state dimension of 1024. The input to that model was a sequence  $CM_1^N=(CM_1, CM_2\dots, CM_N)$, where $CM_i$, the output of ATA for the $i$th sentence of a post, is a 436 dimensional vector and $N$ is the sequence length. To predict the labels of a sequence, we concatenate the last hidden states of the last layer in forward ($\overrightarrow{h_n}$) and backward directions ($ \overleftarrow{h_n}$). The result vector of concatenation $h_n = [\overrightarrow{h_n}|\overleftarrow{h_n}]$ is then transformed through a 2-layer feedforward neural network, whose activation function is Rectifier Linear Unit \cite{agarap2018deep}. The output of this is then passed to a Fully Connected (FC) layer with ReLu activation function and dropout of 0.2 and it is fed to a final FC layer. The output is passed through sigmoid function and finally a threshold is used to determine the labels. We trained these models for 500 epochs, and saved the model that performs best on validation set, with a batch size of 256 and a sequence length of 10. The fourth model (Hybrid) is a hybrid classification model that integrates (i) a pretrained RoBERTa model whose output is passed through a BiLSTM layer and a subsequent FC layer with (ii) a BiLSTM network of linguistic features of the text with a subsequent FC layer. The FC layers of both components take as input the concatenation of last hidden states of the last BiLSTM layer in forward and backward direction. We concatenated the outputs of these components before finally feeding them into a final FC layer with a sigmoid activation function. Specifically, the component with the pretrained RoBERTa model comprised a 2-layer BiLSTM with 256 hidden units and a dropout of 0.2. The component with the with the linguistic features consists of a 3-layer BiLSTM with a hidden size of 512 and a dropout of 0.2. We trained this model for 12 epochs, saving the model with the best performance (F1-Score) on the development set. The optimizer used is AdamW with a learning rate of 2e-5 and a weight decay of 1e-4. Structure diagrams of the model based solely on linguistic features and the hybrid architectures are presented in Figures \ref{fig:strucPsy} and \ref{fig:strucHyb}. In order to reduce the variance of the estimates, we trained all models in a 5-fold CV setup. Reported values represent averages over five runs. The fifth model (Stacking) applied a stacking approach to ensemble all models \cite{wolpert1992stacked}. 

\begin{figure}
    \centering
    \includegraphics[width = 0.45\textwidth]{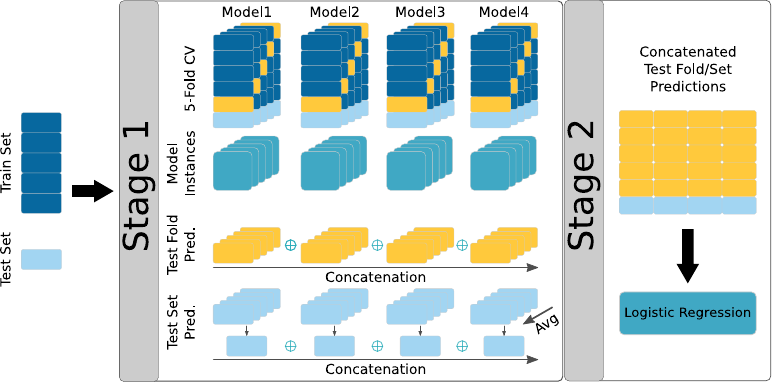}
    \caption{Schematic representation of ensembling by stacking.}
    \label{fig:modelC}
\end{figure}

The training procedure consisted of two stages (see Figure \ref{fig:modelC}). In Stage 1, each of the four models was trained independently using 5-fold cross-validation. For each text sample in the test fold, we obtained a prediction vector from each of the four component models.  These predictions vectors were then concatenated and constituted the input data in a subsequent training stage (Stage 2). The final predictions of the ensemble model were derived from another logistic regression model trained on the concatenated prediction vectors from Stage 1. To perform inference on the test set, the predictions of all model instances trained in Phase 1 were taken and averaged by model to serve as input to Phase 2 after concatenation. All hyperparameters for the training of each of the ensembled models were selected as specified above.

\subsection{Feature ablation} 

To assess the relative importance of the feature groups in predicting six mental health conditions, we used Submodular Pick Lime (SP-LIME; \cite{ribeiro2016should}). SP-LIME is a method to construct a global explanation of a model by aggregating the weights of linear models, that locally approximate the original model. To this end, we first constructed local explanations using LIME. Analogous to super-pixels for images, we categorized our features into eight groups (see section 3.2). We used binary vectors $z\in\{0,1\}^{d}$ to denote the absence and presence of feature groups in the perturbed data samples, where $d$ is the number of feature groups. Here, `absent' means that all values of the features in the feature group are set to 0, and `present' means that their values are retained. For simplicity, a linear regression model was chosen as the local explanatory model. An exponential kernel function with Hamming distance and kernel width $\sigma=0.75\sqrt{d}$ was used to assign different weights to each perturbed data sample. After constructing their local explanation for each data sample in the original dataset, the matrix $W\in\mathbb{R}^{n\times d}$ was obtained, where $n$ is the number of data samples in the original dataset and $W_{ij}$ is the $j$th coefficient of the fitted linear regression model to explain data sample $x_i$. The global importance score of the SP-LIME for feature $j$ can then be derived by: $I_j = \sqrt{\sum_{i=1}^n |W_{ij}|}$

\section{Results and Discussion} \label{results}

\begin{table*}[htb!]
\centering
\caption{Results of the multiclass classification. All numbers represent F1 scores averaged across 5 folds.}
\label{tab:RES}
\adjustbox{max width=0.85\textwidth}{
\begin{tabular}{|l|lllllll|l|}
\hline
                & \multicolumn{7}{c|}{\textbf{Mental Health Condition}}             &                  \\ \cline{2-8}
\textbf{Models} & Depression & Anxiety & Bipolar & ADHD  & Stress & PTSD  & Control & \textbf{Average} \\ \hline
BERT & 17.40 & 15.20 & 19.80 & 5.80 & 71.20 & 7.60 & 48.00 & 28.00 \\
RoBERTa    & 27.48      & 12.83   & 3.46    & 17.88 & 76.22  & 1.46  & 52.85   & 27.45             \\
Psyling-BiLSTM & 19.40      & 15.80   & 9.60    & 14.60 & 51.80  & 4.00  & 36.60   & 22.20            \\
Hybrid          & 18.40      & 17.00   & 11.80   & 19.40 & 77.00  & 14.00 & 50.60   & 29.80            \\
Stacking & 27.23 & 18.55 & 18.21 & 24.84 & 76.61 & 0.96 & 53.58 & 31.40 \\
\hline
\end{tabular}
}
\vspace{-3mm}
\end{table*}

Table \ref{tab:RES} gives an overview of the results of the five multiclass classification models described in Section 3.2 Our overall best-performing model (Stacking) achieved a macro F1 score of 31.4\%, corresponding to an increase in performance of +3.4\% F1 over the BERT baseline and +3.95\% F1 over the RoBERTa baseline. In terms of class-wise performance, the highest prediction accuracy was achieved in the detection of stress with a maximum average F1 score of 77\%. The second highest prediction accuracy was achieved for the control class with a maximum average F1 score of 53.58\%. The next highest classification accuracies were observed for depression ($27.48\%$ F1) and ADHD  ($24.84\%$ F1).
Anxiety and bipolar exhibited maximum prediction accuracies greater than 18\% F1. Lowest accuracy (14\%) was obtained for PTSD. Our Psyling-BiLSTM-model trained exclusively on within-text distributions of eight feature groups achieved a macro F1 score of 22.20\%, a decrease of -5.8\% F1 from the BERT baseline and -5.25\% F1 from the RoBERTa baseline. Another key finding of our experiments is that mental health state prediction benefits immensely from a hybrid approach: The results show that a hybrid model integrating a RoBERTa-based model with text-internal distributions of eight feature groups outperforms the transformer-based models by +1.8\% (vs. BERT) and +2.35\% (vs. RoBERTa) macro-F1. Moreover, the hybrid model efficiently combined the strengths of the two transformer models (BERT and RoBERTa) and Psyling-BiLSTM, which significantly increased the robustness of the model predictions: Both the transformer-based baseline models and the Psyling-BiLSTM showed below chance performance ($<12.5$ \% F1) for two of the seven classes. The hybrid model compensated for such drawbacks in an effective manner.

\begin{figure}
    \centering
    \includegraphics[width = 0.5\textwidth]{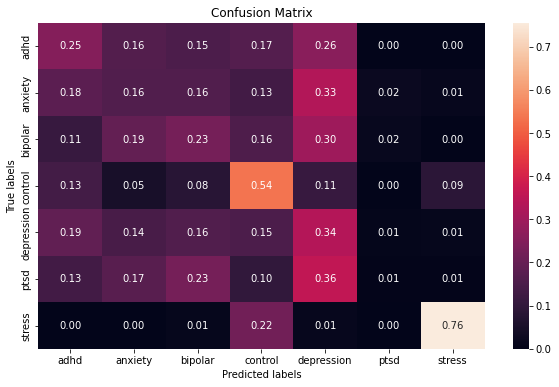}
    \caption{Confusion matrix of the stacking model on multi-class mental health status prediction.}
    \label{fig:cmn_withptsd}
    \vspace{-5mm}
\end{figure}

As for the error analysis, Figure \ref{fig:cmn_withptsd} shows the confusion matrix of our best model (Stacking) normalized over the actual classes (in rows). We found that for five of the seven mental health conditions, the majority of model predictions applied to the correct class (ADHD 25\%, bipolar 23\%, depression 34\%, stress 76\%, control 54\%). Bipolar disorder was frequently misclassified as PTSD (23\%). Anxiety was most often classified as ADHD (18\%), followed by bipolar disorder and correct classification (both 16\%). Depression posts were most frequently confused with ADHD (19\%), bipolar disorder (16\%) and anxiety (14\%). At the same time, depression was by far the most frequently predicted class overall, with an average prediction rate of 24.4\%. 

\begin{table*}
\caption{Results of the feature ablation. Values represents I scores of a feature group in percent. Values in parentheses indicate the rank of a feature groups per MHC.}
\label{tab:importance}
\centering
\adjustbox{max width=0.85\textwidth}{
\begin{tabular}{|l|llllll|}
\hline
\textbf{}              & \multicolumn{6}{c|}{\textbf{Importance}}                                                                                       \\ \cline{2-7} 
\textbf{Feature Group} & \textbf{Depression} & \textbf{Anxiety} & \textbf{Adhd} & \textbf{Bipolar} & \textbf{Stress} & \textbf{Ptsd} \\ \hline
Readability  (N=14)          & 37.06 (1)            & 38.83 (1)          & 34.68 (1)     & 40.1 (1)         & 25.14 (2)       & 41.86 (1)           \\
Reg.-spec. Ngram  (N=25)                & 21.85 (2)           & 21.11 (2)        & 24.02 (2)     & 20.56 (2)        & 21.43 (3)       & 20 (2)              \\
Lexical richness (N=52)               & 15.92 (3)           & 15.17 (3)        & 15.48 (3)      & 14.73 (3)        & 26.15 (1)       & 14.27 (3)             \\
EmoSent   (N=39)             & 12.09 (4)           & 11.98 (4)         & 11.79 (4)     & 11.87 (4)        & 12.18 (4)       & 11.46 (4)            \\
MorphSyn complexity (N=19)               & 8.01 (5)            & 7.94 (5)         & 8.69 (5)      & 7.81 (5)         & 9.47 (5)         & 7.7 (5)             \\
LIWC       (N=61)            & 2.48 (6)            & 2.42 (6)         & 2.61 (6)      & 2.41 (6)         & 2.71 (6)        & 2.29 (6)        \\
General Inquirer    (N=188)           & 1.98 (7)            & 1.94 (7)         & 2.08 (7)      & 1.91 (7)         & 2.22 (7)        & 1.84 (7)             \\
GALC        (N=38)           & 0.62 (8)            & 0.61 (8)         & 0.66 (8)      & 0.6 (8)          & 0.69 (8)        & 0.58 (8)             \\ \hline
\end{tabular}
}
\vspace{-3mm}
\end{table*}
These findings reflect evidence in the psychiatric literature indicating that there is considerable overlap in clinical symptoms and pathophysiological processes and that depressive symptoms may also occur in the context of another psychiatric disorder (e.g., bipolar disorder) \cite{baldwin2002can}. Furthermore, psychiatric data suggest that depressive disorders (i.e., major depressive disorder and dysthymia) are highly comorbid with other common mental disorders \cite{rohde1991comorbidity,gold2020comorbid}. In contrast,  misclassifications in the stress category were almost exclusively controls (22\% of all predictions), indicating that statistical patterns of language use reflecting stress differ from those for diagnosed mental health disorders. Controls were in turn most frequently confused with ADHD (13\% of all predictions). This finding is consistent with the prevalence of overdiagnosis of ADHD in children and adolescents \cite{kazda2019evidence}. Finally, PTSD was correctly classisfied in only 1\% of the cases, and typically misclassified as depression (36\%) or bipolar (23\%). That said, user posts were predicted by the stacking model to be PTSD only 6.5\% (21/320) of the time, suggesting that the classifier is sensitive to the slightly lower frequency of this mental disorder. In view of the model's tendency to avoid predictions for the less populated class, we conducted additional multiclass experiments without the PTSD class to determine how this would affect the overall pattern of findings. The results of these experiments revealed that the exclusion of PTSD yielded a slight improvement in overall classification accuracy, with the improvement over chance increasing from 18.9\% F1 to 23.65\% F1. In regards to rank order, the performances of the models mirror those of the models with PTSD: the hybrid model still outperformed both transformer-based models (+3.6\% F1 over BERT and +3.37\% F1 over RoBERTa) and the stacked generalization still yielded  highest classification accuracy (+2.05\% F1 over the hybrid model). The general patterns of misclassification remained the same (for further details, see Table \ref{tab:RESwoPTSD} in the appendix).

The results of the feature ablation experiments are presented in Table \ref{tab:importance}. We found that the three most important feature groups across all six mental health conditions are rather general in nature: Readability, lexical richness, and register-specific n-gram frequencies. In comparison, the feature groups representing closed vocabulary approaches (EmoSent, LIWC, General Inquirer, GALC), which have been prominently used in previous work on health text mining, play a minor role. This is particularly striking given that these groups comprise a much greater number of features that have repeatedly been identified as mental health signals (see, e.g., Resnik et al., 2013; Alvarez-Conrad et al., 2001; Tausczik and Pennebaker, 2010, Coppersmith et al., 2014). It is noteworthy that the ranking of the three most important feature groups is consistent across all five mental disorders assessed, with readability features being the most important group. In contrast, stress is strongly associated with features of lexical richness, which includes measures of lexical sophistication, variety, and density. Taken together, these results suggest that research in health text mining and automatic prediction of mental health conditions should move beyond lexicon-based feature groups and place a greater emphasis on more general text features.

\section{Conclusion and Outlook} \label{conclusion}

In this paper, we reported on multiclass classification experiments aimed at predicting six mental health conditions from Reddit social media posts. We explored and compared the performance of hybrid and ensemble models leveraging transformer-based architectures (BERT and RoBERTa) and BiLSTM networks trained on within-text distributions of a diverse set of linguistic features. Our results show that the proposed hybrid models significantly improve both model robustness and model accuracy compared to transformer-based baseline models. The use of model stacking proved to be an effective technique to further improve model accuracy. Ablation experiments revealed 
that the importance of textual features concerning readability, register-specific n-gram frequency and lexical richness far outweighs the importance of closed vocabulary features. In future work, we intend to perform comprehensive feature analysis based on within-text distribution to identify most distinctive indicators of diverse depressive disorders. We also intend to extend the approach presented here to incorporate features of textual cohesion. In addition, we intend to integrate the proposed approach with data on the behavioral activity of the individual, such as the frequency of posting and the temporal distribution of posting histories.


\FloatBarrier

\bibliography{anthology,custom}
\bibliographystyle{acl_natbib}

\clearpage
\FloatBarrier
\appendix

\onecolumn
\section{Appendix}
\label{sec:appendix}

\FloatBarrier
\begin{table*}[h!]
  \centering
  \setlength{\tabcolsep}{2pt}
  \caption{Overview of the 436 features investigated in the work.}
  \label{tab:features}
    \begin{tabular}{|l|c|l|l|}
		\hline
		Feature group & Number & Features & Example/Description \\
		& of features &  & \\
		\hline
		Morpho-syntactic & 19    &MLC&Mean length of clause (words)\\

		&&MLS&Mean length of sentence (words)\\
		&&MLT&Mean length of T-unit (words)\\
		&&C/S&Clauses per sentence\\
		&&C/T&Clauses per T-unit\\
		&&DepC/C&Dependent clauses per clause\\
		&&T/S&T-units per sentence\\
		&&CompT/T&Complex T-unit per T-unit\\
		&&DepC/T&Dependent Clause per T-unit\\
		&&CoordP/C&Coordinate phrases per clause\\
		&&CoordP/T&Coordinate phrases per T-unit\\
		&&NP.PostMod&NP post-mod (word)\\
		&&NP.PreMod&NP pre-mod (word)\\
		&&CompN/C&Complex nominals per clause\\
		&&CompN/T&Complex nominals per T-unit\\
		&&VP/T&Verb phrases per T-unit\\
            && BaseKolDef& Kolmogorov Complexity\\
            && MorKolDef& Morphological Kolmogorov Complexity\\
            &&SynKolDef & Syntactic Kolmogorov Complexity\\
		\hline
		Lexical richness & 52    & MLWc & Mean length per word (characters) \\
		&&MLWs&Mean length per word (sylables)\\
		&&LD&Lexical density\\
		&&NDW&Number of different words\\
		&&CNDW&NDW corrected by Number of words\\
		&&TTR&Type-Token Ration (TTR)\\
		&&cTTR&Corrected TTR\\
		&&rTTR&Root TTR\\
		&&AFL&Sequences Academic Formula List\\
		&&ANC& LS (ANC) (top 2000)\\
		&&BNC&LS (BNC) (top 2000)\\
		&&NAWL&LS New Academic Word List\\
		&&NGSL&LS (General Service List) \\
		&&NonStopWordsRate& Ratio of words in NLTK non-stopword list\\
    &     &  WordPrevalence & See \citet{brysbaert2019word} \\
		&& Prevalence & Word prevalence list\\
		&&&incl. 35 categories \\
		&&&(\citet{johns2020estimating})\\
		&&AoA-mean&avg. age of acquisition \\
		&&&(\citet{kuperman2012age})\\
		&&AoA-max&max. age of acquisition\\
	\hline
	\end{tabular}%

\end{table*}

\begin{table*}
    \centering
    
  \begin{tabular}{|l|c|l|l|}
		\hline
		(continued)& &  & \\
		\hline
		Register-based  &   25  & Spoken ($n\in [1,5]$) & Frequencies of uni-, bi-\\
	N-gram	&       & Fiction ($n\in [1,5]$) & tri-, four-, five-grams\\
		&       & Magazine ($n\in [1,5]$) &  from the five sub-components \\
		&       & News ($n\in [1,5]$) & (genres) of the COCA,\\
		&       & Academic ($n\in [1,5]$) &  see \citet{davies2008corpus}\\
		\hline
		Readability & 14     & ARI  &  Automated Readability Index \\
		&       & ColemanLiau & Coleman-Liau Index \\
		&       & DaleChall & Dale-Chall readability score\\
		&& FleshKincaidGradeLevel&Flesch-Kincaid Grade Level\\ 
		&& FleshKincaidReadingEase&Flesch Reading Ease score\\
		&& Fry-x&x coord. on Fry Readability Graph\\
		&& Fry-y&y coord. on Fry Readability Graph\\
		&&Lix&Lix readability score\\
		&&SMOG&Simple Measure of Gobbledygook\\
		&&GunningFog&Gunning Fog Index readability score\\
		&&DaleChallPSK&Powers-Sumner-Kearl Variation of \\
		&&&the Dale and Chall Readability score\\
		&&FORCAST&FORCAST readability score\\
		&&Rix&Rix readability score\\
		&&Spache&Spache readability score\\
		
		\hline
\textbf{Lexicons:}  & 325    &  & \\
 EmoSent &   39  & ANEW-Emo lexicons & \citep{stevenson2007characterization}\\
   &     & Affective Norms for English Words & \citep{bradley1999affective}\\
    &     &DepecheMood++  & \citep{araque2019depechemood++} \\
    &     & NRC Word-Emotion Association  & \citep{mohammad2013crowdsourcing} \\
         &     & NRC Valence, Arousal, and Dominance   &  \citep{mohammad2018obtaining}\\
          &     & SenticNet &\citep{cambria2010senticnet}  \\
           &     & Sentiment140 & \citep{MohammadKZ2013} \\
    GALC & 38    & Geneva Affect Label Coder   & \citep{scherer2005emotions} \\

     LIWC  & 61    & LIWC & \citep{pennebaker2001linguistic} \\
       Inquirer      &  188   & General Inquirer  & \citep{stone1966general} \\
    \hline
	\end{tabular}%
\end{table*}

\begin{table*}
\centering
\caption{Results of the multiclass classification of MHCs (without PTSD).}
\label{tab:RESwoPTSD}
\begin{tabular}{|l|llllll|l|}
\hline
                & \multicolumn{6}{c|}{\textbf{Mental Health Condition}}             &                  \\ \cline{2-7}
\textbf{Models} & Depression & Anxiety & Bipolar & ADHD  & Stress  & Control & \textbf{Average} \\ \hline
BERT & 4.36 & 29.12 & 3.47 & 28.88 & 77.37 & 52.22 & 32.2 \\
RoBERTa & 8.07 & 6.40 & 26.00 & 18.84 & 82.8 & 52.26 & 32.43 \\
Psyling-BiLSTM & 11.48 & 6.88 & 11.43 & 21.25 & 59.00 & 38.32 & 24.84\\
Hybrid & 20.80 & 16.00 & 14.2 & 26.8 & 81.60 & 52.6 & 35.80\\
Model Stacking & 21.93 & 18.96 & 21.93 & 19.10 & 83.14 & 55.22 & 37.85 \\
\hline
\end{tabular}
\end{table*}

\end{document}